\pdfoutput=1

\documentclass[11pt]{article}

\usepackage[preprint]{coling}

\usepackage{times}
\usepackage{latexsym}

\usepackage[greek, english]{babel}  

\usepackage[T1]{fontenc}

\usepackage[utf8]{inputenc}

\usepackage{microtype}

\usepackage{inconsolata}

\usepackage{graphicx}

\usepackage{pgfplots}
\pgfplotsset{compat=1.18} 
\usepackage{pgfplotstable} 

\usepackage{amsmath}

\usepackage{hyperref}

\usepackage{ifpdf}

\title{Automatic Speech Recognition for Greek Medical Dictation%
\ifpdf
\thanks{This preprint presents a shortened version of the undergraduate thesis of Vardis Georgilas (\url{http://nlp.cs.aueb.gr/theses/bsc_thesis_georgilas.pdf}), completed at the Department of Informatics, Athens University of Economics and Business.}%
\fi
}

\author{
Vardis Georgilas$^{1}$, \hspace{0.5mm} 
Themos Stafylakis$^{1,2,3}$ \\
$^{1}$Department of Informatics, Athens University of Economics and Business, Greece\\
$^{2}$Omilia Natural Language Solutions Ltd., Greece\\
$^{3}$Archimedes/Athena RC, Greece
}

\begin{document}
\maketitle
\begin{abstract}
Medical dictation systems are essential tools in modern healthcare, enabling accurate and efficient conversion of speech into written medical documentation. The main objective of this paper is to create a domain-specific system for Greek medical speech transcriptions. The ultimate goal is to assist healthcare professionals by reducing the overload of manual documentation and improving workflow efficiency. Towards this goal, we develop a system that combines automatic speech recognition techniques with text correction model, allowing better handling of domain-specific terminology and linguistic variations in Greek. Our approach leverages both acoustic and textual modeling to create more realistic and reliable transcriptions. We focused on adapting existing language and speech technologies to the Greek medical context, addressing challenges such as complex medical terminology and linguistic inconsistencies. Through domain-specific fine-tuning, our system achieves more accurate and coherent transcriptions, contributing to the development of practical language technologies for the Greek healthcare sector.

\end{abstract}

\section{Introduction}

Medical dictation plays a pivotal role in modern healthcare workflows \cite{alhadidi2017dictation}. Precise and timely documentation of medical information enhances diagnostic accuracy, ensures continuity of care, and supports legal protection. Traditional documentation is time-consuming for healthcare professionals, requiring them to spend more time in each case. Dictation systems provide a fast and natural alternative to manual data entry, reducing documentation workload and improving healthcare professionals efficiency. 

Greek medical dictation is under-resourced compared to English systems. Existing speech recognition systems perform poorly in Greek medical domain due to complex domain-specific terminology, the linguistic characteristics of the Greek language and possibly due to variations in individual speech patterns. 

We develop a system that combines state-of-the-art automatic speech recognition techniques with domain adapted language model for Greek medical dictation. Our ASR component uses the pre-trained Whisper model \cite{radford2023whisper}, adapted to Greek. Additionaly, a Greek version of GPT-2 \cite{radford2019language} language model, fine-tuned with Greek medical text, is used to evaluate and select the best transcription hypothesis from the multiple candidates generated by the ASR. This approach integrates both acoustic and linguistic information to improve transcription accuracy and and better handle specialized medical terminology.

\section{Background and Related Work}

Early ASR systems were based on statistical models such as Hidden Markov Models (HMMs) with Gaussian Mixture Models (GMMs) for acoustic modeling \cite{Bahl83}. These approaches relied on strong independent assumptions and struggled with complex phonetic patterns. Neural networks, especially RNNs and LSTMs, improved performance by modeling long-term dependencies in speech. In addition, the introduction of the Transformer architecture changed the approach to sequence processing, further advanced ASR with self-attention, enabling efficient end-to-end architectures \cite{Vaswani2017}.

Modern speech recognition systems tend to combine acoustic, pronunciation, and language models into a single unified network that can be trained end-to-end. This approach simplifies the pipeline, reduces errors, and improves accuracy. Recent advances in self-supervised learning allow models to be pre-trained on large amounts of unlabeled speech and then fine-tuned on smaller, specialized datasets \cite{Baevski20}, which is especially important for low-resource languages such as Greek.

Medical speech introduces unique challenges, including specialized terminology, abbreviations, complex sentence structures, and context-specific phrases. Clinical environments often involve background noise, multiple speakers, and time constrained communication. General-purpose ASR systems frequently fail in this setting, and the problem is compounded in low-resource languages. Accurate transcription is critical for proper clinical documentation, workflow efficiency, and reducing the risk of medical errors \cite{ng2025ai}.

Language Models play a crucial role in the post-processing of texts generated by automatic speech recognition systems. After the initial conversion from speech to text, the resulting text often contains errors, omissions or inconsistencies. These models help with the correction of those errors, resulting in a more coherent and well-structured transcription. Transformer-based models, like GPT-2 \cite{radford2019language}, have the capability to understand the context of each sentence and provide improvements that make the text more natural and understandable. Domain-specific LMs can be trained using task-relevant text to expand available resources, which has been shown to reduce perplexity and improve transcription quality \cite{Jha2021}. By training these models on large volumes of text, they can identify incorrect sentences or unusual expressions, and improve the flow and clarity. Using them for post-processing is especially important in this task, where accurate transcription and clear understanding of texts are critical.

Fine-tuning pretrained models to specific domains is essential for improving performance without requiring massive datasets or training from scratch. Transfer learning allows models trained on general speech or text corpora to adapt to specialized tasks, such as medical dictation, by learning domain-specific terminology and phrasing. Parameter-efficient methods, such as Low-Rank Adaptation (LoRA)\cite{hulora}, enable large models to be adapted with minimal computational overhead. Instead of updating all parameters, LoRA introduces small learnable matrices that approximate the necessary weight updates while keeping the original model frozen. In this work, we use LoRA to fine-tune each model, allowing the system to handle specialized terminology and domain-specific phrasing, ultimately improving transcription accuracy and reliability.

\section{System Design} 

\begin{figure*}[!t]
    \centering
    \includegraphics[width=\textwidth]{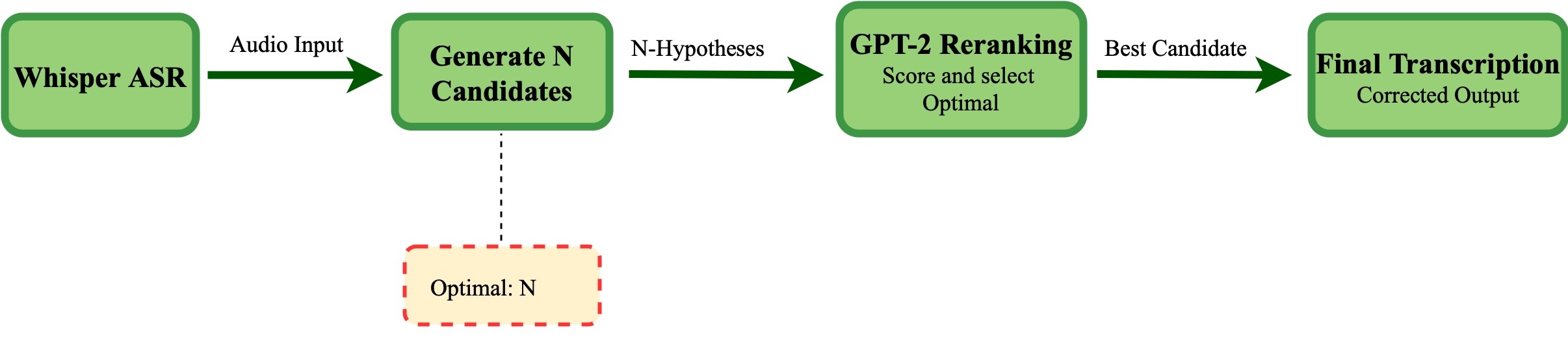}
    \caption{Pipeline for generating and refining speech transcriptions}
    \label{fig:pipeline}
\end{figure*}

This section presents the design and implementation of the Automatic Speech Recognition (ASR) system for Greek medical dictation. The core of this work involves adapting OpenAI's Whisper model to the specific features of the Greek language through a controlled fine-tuning process. A key aspect of our methodology is the comparative analysis of three different sizes of the Whisper model, small, medium, and large-v2. To further enhance transcription quality, we integrated a re-ranking mechanism based on a fine-tuned GPT-2 model, which was used to select the most contextually appropriate transcription among Whisper’s alternatives.

The foundation of our system is the Whisper model \cite{radford2023whisper}, a state-of-the-art ASR model using an encoder-decoder architecture. Whisper has been trained on a large variety of labeled speech data, enabling robust performance across multiple languages and noisy environments. The model converts raw audio into log-Mel spectrogram features, which are processed by convolutional layers and transformer encoder blocks, and decoded into text using cross-attention and a transformer decoder \cite{VasAttention}.

We examine three pre-trained versions of Whisper: small, medium, and large-v2. The primary motive for training three distinct models was to analyze the trade-off between performance and computational cost. While larger models like large-v2 are expected to have higher accuracy due to their increased parameters and greater representational capacity, they are also more computationally expensive and harder to deploy in resource-constrained environments. By fine-tuning and evaluating all three models we can determine the optimal one that meets our desired accuracy benchmarks while being practical and efficient for real-world medical dictation.  

To adapt Whisper model to the Greek language we applied a fine-tuning methodology to each of the three models. A diverse set of different Greek speech datasets was aggregate to create a robust corpus of data, containing speech data with different domains, with varying acoustic environments, and from multiple speakers. The need for such diversity is motivated by the challenges posed by Greek dialectal variation \cite{vakirtzian2024speech}. To make fine-tuning more efficient, Low-Rank Adaptation (LoRA) \cite{hulora} was applied to reduce the number of trainable parameters. In this way, only about $2\%$ of the full model's trainable parameters are trainable, allowing efficient fine-tuning under limited resources.

\begin{table}[h!]
\centering
\small
\begin{tabular}{l c c}
\hline
Model & Params (B) & Trainable Params (M) \\
\hline
Whisper Small    & 0.249 & $\sim$7.1 (2.8\%)    \\
Whisper Medium   & 0.783 & $\sim$18.9 (2.4\%)  \\
Whisper Large-v2 & 1.574 & $\sim$31.5 (2\%) \\
\hline
\end{tabular}
\caption{Trainable Parameters Comparison of Whisper Models Using LoRA}
\end{table}

To improve transcription quality, the Greek GPT-2 model \cite{radford2019language} was used to re-rank N-best hypotheses generated by Whisper. GPT-2 uses a unidirectional transformer decoder with masked multi-head self-attention to model dependencies from left to right \cite{VasAttention}. Our variant, fine-tuned for Greek text, consists of 12 decoder layers, each with 12 attention heads and a hidden size of 768. This model was further fine-tuned on a domain-specific medical text corpus using LoRA \cite{hulora}. Attention and projection layers were adapted with low-rank matrices, updating only 1.29\% of the total parameters. The fine-tuned model was evaluated using perplexity, showing improved confidence in next-token predictions and better alignment with medical terminology.

Our final pipeline integrates both Whisper and Greek GPT-2 (Figure~\ref{fig:pipeline}). Whisper generates a set of N candidate hypotheses, representing possible transcriptions. It converts raw audio waveforms into text using its pre-trained encoder-decoder architecture and produces multiple transcription variants to capture potential ambiguities in the audio. The language model evaluates the N candidate hypotheses. Each candidate is scored based on grammatical correctness, contextual relevance, and semantic coherence. Through this re-ranking process, the most accurate candidate is selected, ensuring that the final transcription aligns well with Greek language conventions. A critical design choice is selecting the optimal value of N, balancing computational load and the diversity of transcription options, enabling GPT-2 to effectively re-rank and select the best candidate. The optimal value of N was determined through empirical testing, ensuring robust performance for Greek audio inputs.

\section{Data}

This work uses several Greek audio datasets used to fine-tune Whisper and a medical text corpus for adapting the Greek GPT-2. Each dataset was assembled with the needs of its respective task in mind, aiming to support effective training, reliable evaluation, and accurate representation of Greek linguistic features.

For Whisper, we fine-tuned the model on a composite dataset of Greek speech audio paired with transcriptions. This dataset combined three publicly available sources to ensure a variety of speakers, accents, and acoustic conditions. The Mosel dataset \cite{gaido2024mosel950000hoursspeech} contains European Parliament recordings in Greek. Some recordings, however, had missing timestamps or misaligned segments. To address this, we curated a filtered dataset of well-aligned audio-transcription pairs for model training \footnote{\url{https://huggingface.co/datasets/Vardis/Greek_Mosel}}. Mozilla Common Voice 11.0 \cite{ardila2020commonvoicemassivelymultilingualspeech} contributed crowd-sourced recordings from volunteers across different accents and environments. Google FLEURS \cite{conneau2022fleursfewshotlearningevaluation} provided read-aloud sentences from multiple domains, complementing the dataset. The final dataset contains approximately 49 hours of speech, covering diverse domains, speaker variability, and acoustic conditions, enabling robust model fine-tuning.

The Greek GPT-2 model was trained on a custom medical text dataset containing 20,430 samples from multiple sources \footnote{\url{https://huggingface.co/datasets/Vardis/Greek\_Medical\_Text}}. Medical e-books provided detailed clinical terminology covering diagnostics, procedures, and patient care \cite{iatrakis:2015, sfikakis:2015, tsipouras:2015}. The QTLP Greek CC Corpus for the medical domain \cite{athena2015_qtlp_gr} added web documents automatically classified as medical, including reference materials, news, discussions, and other genres. Dialogues from medical podcasts, collected from \texttt{istorima.org}, introduced conversational medical language, enriching the dataset with contextual and informal expressions. This corpus allows GPT-2 to rank candidate sentences produced by Whisper based on perplexity, improving transcription selection and alignment with domain-specific terminology.

\section{Evaluation and Results}

We evaluated Whisper and GPT-2, focusing on the final Whisper-GPT-2 pipeline. Each model was evaluated on its respective test dataset to assess its effectiveness in handling Greek medical speech tasks. The Whisper model was fine-tuned in three configurations (Small, Medium, Large-v2) on a composite dataset comprising Greek speech data \cite{gaido2024mosel950000hoursspeech, ardila2020commonvoicemassivelymultilingualspeech, conneau2022fleursfewshotlearningevaluation}. The results are summarized in Table~\ref{whisper_results}, reporting the Word Error Rate (WER), normalized Word Error Rate (nWER), Character Error Rate (CER), and BLEU~\cite{papineni-etal-2002-bleu}. Fine-tuning improved performance across all model sizes. Whisper Small reduced WER from 43.62\% to 30.31\%, Whisper Medium from 34.71\% to 19.45\%, and Whisper Large-v2 from 26.41\% to 14.90\%. Similar gains were observed in CER, confirming the effectiveness of fine-tuning, while nWER, which disregards differences in case, punctuation, and spacing, followed the same trends, further demonstrating Whisper's improved adaptation to Greek.

The Greek GPT-2 model was fine-tuned on a domain-specific corpus designed to adapt the model to the medical context. The training set combined medical texts and transcribed speech data. This hybrid approach allowed the model to learn not only specialized vocabulary but also the stylistic and syntactic patterns typical of oral communication, which are highly relevant for correcting ASR outputs. The evaluation was conducted using perplexity on both medical text , speech transcription, and their combination. Across all three evaluation settings, the fine-tuned model consistently outperformed the original pre-trained Greek GPT-2. More specifically, perplexity was substantially reduced on both medical texts and speech data, confirming that the model successfully learned domain-specific terminology as well as the idiomatic patterns of spoken language. The combined results further highlight the overall effectiveness of the fine-tuning strategy (Table~\ref{gpt2_perplexity}).

\begin{table}[h!]
\centering
\small
\begin{tabular}{l@{\hskip 0pt} c@{\hskip 6pt} c@{\hskip 6pt} c}
\hline
\textbf{Dataset} & \textbf{Pre-trained} & \textbf{Fine-tuned} & \textbf{$\Delta$ (\%)}\\
\hline
Medical Texts & \textbf{45.73} & \textbf{35.36} & 22.7 \\
Speech Transcriptions & 103.21 & 67.67 & \textbf{34.4} \\
Combined (All Data) & 53.15 & 39.86 & 25.0 \\
\hline
\end{tabular}
\caption{Perplexity of Greek GPT-2 (Pre-trained vs Fine-tuned) on Medical and Speech Datasets}
\label{gpt2_perplexity}
\end{table}

The Whisper-GPT-2 pipeline was evaluated on the test dataset using WER, nWER, CER, and BLEU metrics, as shown in Table~\ref{whisper_results}. The pipeline incorporates a re-ranking step in which the Whisper model first generates $N$ candidate transcriptions for each audio segment. Subsequently, the fine-tuned GPT-2 model evaluates these candidates and selects the most probable sentence. Experiments with $N=3,5,8$ showed that $N=5$ provides the best balance between transcription quality and computational efficiency. Increasing $N$ gave only marginal gains but significantly increased computation time, while smaller $N$ values limited GPT-2's re-ranking capabilities.

\begin{table*}[!t]
\centering
\begin{tabular}{lcccc}
\hline
\textbf{Model / Pipeline} & \textbf{WER (\%)} & \textbf{nWER (\%)} & \textbf{CER (\%)} & \textbf{BLEU (\%)} \\
\hline
Original Whisper Small              & 43.62 & 36.69 & 21.61 & 72.61 \\
Fine-tuned Whisper Small            & 30.31 & 26.54 & 13.28 & 82.35 \\
Fine-tuned Whisper Small + GPT-2    & 27.38 & 23.57 & 11.80 & 84.17 \\
\hline
Original Whisper Medium             & 34.71 & 27.21 & 19.30 & 79.14 \\
Fine-tuned Whisper Medium           & 19.45 & 16.17 & 8.96  & 88.93 \\
Fine-tuned Whisper Medium + GPT-2   & 18.23 & 14.86 & \textbf{8.35} & 89.60 \\
\hline
Original Whisper Large-v2            & 26.41 & 18.86 & 14.55 & 82.49 \\
Fine-tuned Whisper Large-v2          & 14.90 & 12.06 & 8.45  & 92.03 \\
Fine-tuned Whisper Large-v2 + GPT-2  & \textbf{14.69} & \textbf{11.98} & 8.66  & \textbf{92.06} \\
\hline
\end{tabular}
\caption{Progression of Whisper models (Original, Fine-tuned, and Fine-tuned + GPT-2 reranking) for Greek ASR.}
\label{whisper_results}
\end{table*}

Re-ranking consistently improves performance across all tested Whisper model sizes. The WER reduction is approximately 9.66\% for the Whisper Small, 6.27\% for the Whisper Medium, and 1.41\% for the Whisper Large-v2. CER and BLEU scores show corresponding gains, highlighting that re-ranking enhances both word-level accuracy and overall sentence quality. Compared to the original models, the full pipeline achieves WER reductions of 37.23\% for Small, 47.45\% for  Medium, and 44.38\% for Large-v2.

Choosing the optimal model depends on deployment constraints. While Whisper Large-v2 achieves the lowest WER, it requires significantly more computational resources, making it less practical for routine deployment. Whisper Medium offers strong performance with lower WER and CER while being faster, making it more practical for real-world applications.

These findings highlight the effectiveness of combining a strong pre-trained ASR model with a domain-adapted language model. The re-ranking stage is very important especially for medical dictation, where even small improvements in transcription accuracy can be vital for understanding specialized terminology and avoiding critical misinterpretations. Overall, this demonstrates that the Whisper-GPT-2 pipeline is an effective approach for improving transcription accuracy and producing higher quality outputs in Greek medical dictation.

\section{Conclusion}

This paper presented an Automatic Speech Recognition (ASR) pipeline tailored to Greek medical dictation, integrating fine-tuned Whisper models with a domain-adapted Greek GPT-2 language model for re-ranking. Through fine-tuning, we achieved substantial reductions in Word Error Rate (WER) and Character Error Rate (CER), while the re-ranking step provided consistent gains across all model sizes. A key contribution of this work is the curation of a high-quality Greek speech-to-text dataset, which addresses issues in existing resources and enables reproducibility.

Beyond quantitative improvements, the proposed pipeline demonstrates the feasibility of combining state-of-the-art ASR with domain-specific language modeling to support the challenging task of Greek medical transcription. By improving accuracy in handling specialized terminology, homophones, and spoken-language variability, this work moves toward reducing the documentation burden for healthcare professionals.

For future work, we aim to incorporate authentic Greek medical speech data, such as doctor–patient interactions and clinical dictations, and explore real-time deployment on GPU-backed systems. These steps will further enhance robustness and bring the system closer to practical adoption in clinical settings.

\bibliography{custom}

\begin{thebibliography}{19}
\providecommand{\natexlab}[1]{#1}

\bibitem[{ath(2015)}]{athena2015_qtlp_gr}
 2015.
\newblock \href {http://hdl.handle.net/11500/ATHENA-0000-0000-2457-6}
  {English-greek qtlp text corpus for the medical domain}.
\newblock Version 1.0.0, dataset (text corpus).

\bibitem[{Al~Hadidi et~al.(2017)Al~Hadidi, Upadhaya, Shastri, and
  Alamarat}]{alhadidi2017dictation}
S.~Al~Hadidi, S.~Upadhaya, R.~Shastri, and Z.~Alamarat. 2017.
\newblock \href {https://doi.org/10.1080/20009666.2017.1379852} {Use of
  dictation as a tool to decrease documentation errors in electronic health
  records}.
\newblock \emph{J Community Hosp Intern Med Perspect}, 7(5):282--286.

\bibitem[{Ardila et~al.(2020)Ardila, Branson, Davis, Henretty, Kohler, Meyer,
  Morais, Saunders, Tyers, and
  Weber}]{ardila2020commonvoicemassivelymultilingualspeech}
Rosana Ardila, Megan Branson, Kelly Davis, Michael Henretty, Michael Kohler,
  Josh Meyer, Reuben Morais, Lindsay Saunders, Francis~M. Tyers, and Gregor
  Weber. 2020.
\newblock \href {https://arxiv.org/abs/1912.06670} {Common voice: A
  massively-multilingual speech corpus}.
\newblock \emph{Preprint}, arXiv:1912.06670.

\bibitem[{Baevski et~al.(2020)Baevski, Zhou, Mohamed, and Auli}]{Baevski20}
Alexei Baevski, Henry Zhou, Abdelrahman Mohamed, and Michael Auli. 2020.
\newblock \href {https://arxiv.org/abs/2006.11477} {wav2vec 2.0: {A} framework
  for self-supervised learning of speech representations}.
\newblock \emph{CoRR}, abs/2006.11477.

\bibitem[{Bahl et~al.(1983)Bahl, Jelinek, and Mercer}]{Bahl83}
Lalit Bahl, Frederick Jelinek, and Robert Mercer. 1983.
\newblock \href {https://doi.org/10.1109/TPAMI.1983.4767370} {A maximum
  likelihood approach to continuous speech recognition}.
\newblock \emph{Pattern Analysis and Machine Intelligence, IEEE Transactions
  on}, PAMI-5:179 -- 190.

\bibitem[{Conneau et~al.(2022)Conneau, Ma, Khanuja, Zhang, Axelrod, Dalmia,
  Riesa, Rivera, and Bapna}]{conneau2022fleursfewshotlearningevaluation}
Alexis Conneau, Min Ma, Simran Khanuja, Yu~Zhang, Vera Axelrod, Siddharth
  Dalmia, Jason Riesa, Clara Rivera, and Ankur Bapna. 2022.
\newblock \href {https://arxiv.org/abs/2205.12446} {Fleurs: Few-shot learning
  evaluation of universal representations of speech}.
\newblock \emph{Preprint}, arXiv:2205.12446.

\bibitem[{Gaido et~al.(2024)Gaido, Papi, Bentivogli, Brutti, Cettolo, Gretter,
  Matassoni, Nabih, and Negri}]{gaido2024mosel950000hoursspeech}
Marco Gaido, Sara Papi, Luisa Bentivogli, Alessio Brutti, Mauro Cettolo,
  Roberto Gretter, Marco Matassoni, Mohamed Nabih, and Matteo Negri. 2024.
\newblock \href {https://arxiv.org/abs/2410.01036} {Mosel: 950,000 hours of
  speech data for open-source speech foundation model training on eu
  languages}.
\newblock \emph{Preprint}, arXiv:2410.01036.

\bibitem[{Hu et~al.(2021)Hu, Shen, Wallis, Allen{-}Zhu, Li, Wang, and
  Chen}]{hulora}
Edward~J. Hu, Yelong Shen, Phillip Wallis, Zeyuan Allen{-}Zhu, Yuanzhi Li,
  Shean Wang, and Weizhu Chen. 2021.
\newblock \href {https://arxiv.org/abs/2106.09685} {Lora: Low-rank adaptation
  of large language models}.
\newblock \emph{CoRR}, abs/2106.09685.

\bibitem[{Iatrakis(2015)}]{iatrakis:2015}
Georgios Iatrakis. 2015.
\newblock \href
  {https://www.ebooks4greeks.gr/gynaikologika-problhmata-kai-lyseis}
  {\emph{Gynecological Problems and Solutions}}.

\bibitem[{Jha(2021)}]{Jha2021}
Saurav Jha. 2021.
\newblock \href {https://arxiv.org/abs/2110.10261} {Learning domain specific
  language models for automatic speech recognition through machine
  translation}.
\newblock \emph{CoRR}, abs/2110.10261.

\bibitem[{Ng et~al.(2025)Ng, Wang, Zhou, Zhou, Goh, Sim, Tan, Goh, and
  Ng}]{ng2025ai}
JJW Ng, E~Wang, X~Zhou, KX~Zhou, CXL Goh, GZN Sim, HK~Tan, SSN Goh, and QX~Ng.
  2025.
\newblock \href {https://doi.org/10.1186/s12911-025-03061-0} {Evaluating the
  performance of artificial intelligence-based speech recognition for clinical
  documentation: a systematic review}.
\newblock \emph{BMC Med Inform Decis Mak}, 25(1).

\bibitem[{Papineni et~al.(2002)Papineni, Roukos, Ward, and
  Zhu}]{papineni-etal-2002-bleu}
Kishore Papineni, Salim Roukos, Todd Ward, and Wei-Jing Zhu. 2002.
\newblock \href {https://doi.org/10.3115/1073083.1073135} {{B}leu: a method for
  automatic evaluation of machine translation}.
\newblock In \emph{Proceedings of the 40th Annual Meeting of the Association
  for Computational Linguistics}, pages 311--318, Philadelphia, Pennsylvania,
  USA. Association for Computational Linguistics.

\bibitem[{Radford et~al.(2023)Radford, Kim, Xu, Brockman, McLeavey, Welinder,
  Sutskever, and Zaremba}]{radford2023whisper}
Alec Radford, Jong~Wook Kim, Tao Xu, Greg Brockman, Christine McLeavey, Peter
  Welinder, Ilya Sutskever, and Wojciech Zaremba. 2023.
\newblock \href {https://arxiv.org/abs/2212.04356} {Robust speech recognition
  via large-scale weak supervision}.
\newblock In \emph{Proceedings of the 40th International Conference on Machine
  Learning (ICML)}.

\bibitem[{Radford et~al.(2019)Radford, Wu, Child, Luan, Amodei, and
  Sutskever}]{radford2019language}
Alec Radford, Jeffrey Wu, Rewon Child, David Luan, Dario Amodei, and Ilya
  Sutskever. 2019.
\newblock \href
  {https://cdn.openai.com/better-language-models/language_models_are_unsupervised_multitask_learners.pdf}
  {Language models are unsupervised multitask learners}.
\newblock OpenAI Blog.
\newblock Technical report.

\bibitem[{Sfikakis et~al.(2015)Sfikakis, Kokkinos, Kyrtsoni, Makrylakis,
  Boletis, Papatheodoridis, Tentolouris, Psychogiou, Daikos, and
  Vlachogiannakos}]{sfikakis:2015}
P.~Sfikakis, A.~Kokkinos, M.Ch. Kyrtsoni, K.~Makrylakis, I.~Boletis,
  G.~Papatheodoridis, N.~Tentolouris, M.~Psychogiou, G.~Daikos, and
  I.~Vlachogiannakos. 2015.
\newblock \href
  {https://www.ebooks4greeks.gr/askhseis-shmeiologias-kai-diaforikhs-diagnwstikhs-sthn-pathologia}
  {\emph{Exercises in Semiology and Differential Diagnostics in Pathology}}.

\bibitem[{Tsipouras et~al.(2015)Tsipouras, Giannakeas, Karvounis, and
  Tzallas}]{tsipouras:2015}
Markos Tsipouras, Nikolaos Giannakeas, Evangelos Karvounis, and Alexandros
  Tzallas. 2015.
\newblock \href
  {https://www.ebooks4greeks.gr/iatrikh-plhroforikh-pshfiakh-epeksergasia-bioiatrikwn-shmatwn}
  {\emph{Medical Informatics}}.

\bibitem[{Vakirtzian et~al.(2024)Vakirtzian, Tsoukala, Bompolas, Mouzou,
  Stamou, Paraskevopoulos, Dimakis, Markantonatou, Ralli, and
  Anastasopoulos}]{vakirtzian2024speech}
Socrates Vakirtzian, Chara Tsoukala, Stavros Bompolas, Katerina Mouzou, Vivian
  Stamou, Georgios Paraskevopoulos, Antonios Dimakis, Stella Markantonatou,
  Angela Ralli, and Antonios Anastasopoulos. 2024.
\newblock \href {https://doi.org/10.21437/Interspeech.2024-24432} {Speech
  recognition for greek dialects: A challenging benchmark}.
\newblock In \emph{Proceedings of Interspeech 2024}, pages 3974--3978.

\bibitem[{Vaswani et~al.(2017)Vaswani, Shazeer, Parmar, Uszkoreit, Jones,
  Gomez, Kaiser, and Polosukhin}]{VasAttention}
Ashish Vaswani, Noam Shazeer, Niki Parmar, Jakob Uszkoreit, Llion Jones,
  Aidan~N. Gomez, Lukasz Kaiser, and Illia Polosukhin. 2017.
\newblock \href {https://arxiv.org/abs/1706.03762} {Attention is all you need}.
\newblock \emph{CoRR}, abs/1706.03762.

\bibitem[{Zhang et~al.(2017)Zhang, Pezeshki, Brakel, Zhang, Laurent, Bengio,
  and Courville}]{Vaswani2017}
Ying Zhang, Mohammad Pezeshki, Philemon Brakel, Saizheng Zhang, C{\'{e}}sar
  Laurent, Yoshua Bengio, and Aaron~C. Courville. 2017.
\newblock \href {https://arxiv.org/abs/1701.02720} {Towards end-to-end speech
  recognition with deep convolutional neural networks}.
\newblock \emph{CoRR}, abs/1701.02720.

\end{thebibliography}

\end{document}